\documentclass[preprint,12pt]{elsarticle}
\usepackage{float}
\usepackage{algorithm,algcompatible,amsmath,amssymb,algpseudocode}
\usepackage{multirow}
\usepackage{xcolor}
\usepackage{graphicx}
\usepackage{makecell}
\usepackage{array}
\usepackage[T1]{fontenc}
\usepackage[utf8]{inputenc}
\algnewcommand\INPUT{\item[\textbf{Input:}]}%
\algnewcommand\OUTPUT{\item[\textbf{Output:}]}%

\begin{document}
\begin{frontmatter}



\title{Context-CrackNet: A Context-Aware Framework for Precise Segmentation of Tiny Cracks in Pavement images}

\cortext[cor1]{Corresponding author}

\author[label1]{Blessing Agyei Kyem}
\author[label1]{Joshua Kofi Asamoah}
\author[label1]{Armstrong Aboah\corref{cor1}}

\affiliation[label1]{
    organization = {North Dakota State University}, 
    city         = {Fargo},
    postcode     = {58105}, 
    state        = {North Dakota},
    country      = {United States}
}




\begin{abstract}
The accurate detection and segmentation of pavement distresses, particularly tiny and small cracks, are critical for early intervention and preventive maintenance in transportation infrastructure. Traditional manual inspection methods are labor-intensive and inconsistent, while existing deep learning models struggle with fine-grained segmentation and computational efficiency. To address these challenges, this study proposes Context-CrackNet, a novel encoder-decoder architecture featuring the Region-Focused Enhancement Module (RFEM) and Context-Aware Global Module (CAGM). These innovations enhance the model's ability to capture fine-grained local details and global contextual dependencies, respectively. Context-CrackNet was rigorously evaluated on ten publicly available crack segmentation datasets, covering diverse pavement distress scenarios. The model consistently outperformed 9 state-of-the-art segmentation frameworks, achieving superior performance metrics such as mIoU and Dice score, while maintaining competitive inference efficiency. Ablation studies confirmed the complementary roles of RFEM and CAGM, with notable improvements in mIoU and Dice score when both modules were integrated. Additionally, the model's balance of precision and computational efficiency highlights its potential for real-time deployment in large-scale pavement monitoring systems.
\end{abstract}




\end{frontmatter}



\section{Introduction}
\label{intro}
Transportation infrastructure is essential to modern society, forming the backbone of economic development by enabling the efficient movement of people and goods. Among these infrastructures, road networks are critical, and maintaining their integrity is imperative to ensure public safety and economic efficiency. Pavement distresses such as cracks and potholes which develop on these road networks not only compromise safety but also lead to costly repairs if not promptly detected and addressed. Accurately detecting these distresses remains a significant challenge due to their irregular shapes, varying sizes, diverse surface textures, and environmental factors such as fluctuating lighting conditions and the presence of debris. Traditionally, the detection of these distresses has relied on manual inspections, which are not only time-consuming and labor-intensive but also prone to subjectivity and inconsistency. These limitations underscore the critical need for automated solutions to improve efficiency and accuracy. To address these challenges, researchers have increasingly turned to automated approaches that leverage advanced image processing and machine learning techniques, offering a more robust and scalable alternative to traditional methods. While early image processing approaches were often inadequate due to the complex nature of pavement surfaces, the introduction of deep learning models particularly convolutional neural networks has significantly advanced the field. These models effectively identify and segment pavement distresses by learning hierarchical features and capturing spatial context. Nevertheless, despite these advancements, current models still face significant limitations that hinder their performance, necessitating further refinements to achieve accurate pavement distress segmentation.

Several deep learning approaches have been proposed for pavement distress segmentation. For instance, Wen et al. proposed PDSNet \cite{PDSNET}, an efficient framework achieving an MIoU of 83.7\% on manually collected 2D and 3D private pavement dataset. However, it struggles with small and tiny cracks .  Sarmiento \cite{Sarmiento2021PavementDD} utilized YOLOv4 for detecting and DeepLabv3 for segmenting the pavement distresses. While effective for simpler distresses like delaminations, both models struggle with tiny cracks, scaling, and texture variations, leading to misclassifications and false negatives. Li et al \cite{deeplabv3+} further introduced a variant of DeepLabV3+ with an adaptive probabilistic sampling method and external attention for pavement distress segmentation. It was evaluated on the CRACK500 dataset, achieving a Mean Intersection over Union (MIoU) of 54.91\%.. Tong et al. introduced Evidential Segmentation Transformer (ES-Transformer) \cite{evidential}, combining Dempster–Shafer theory with a transformer backbone for improved segmentation and calibration. Evaluated on the Crack500 dataset, it achieved a Mean Intersection over Union (MIoU) of 59.85\%, demonstrating superior performance. However, the architecture introduced is computationally expensive since the transformer architecture used scales quadratically with the input data. Kyem et al. \cite{kyem2024pavecapmultimodalframeworkcomprehensive} used YOLOv8 and the Segment Anything Model (SAM) for segmentation in their PaveCap framework, utilizing SAM’s zero-shot capability for generating binary masks. However, the method struggled with accurately segmenting mixed or overlapping pavement distresses. Owor et al. also introduced PaveSAM \cite{pavesam}, a zero-shot segmentation model fine-tuned for pavement distresses using bounding box prompts, significantly reducing labeling costs and achieving strong performance. One problem with this model is its inability to segment fine-grained distresses in the pavement images.

Despite the significant progress achieved with deep learning models for pavement distress segmentation, several limitations remain. One significant, yet unresolved challenge in pavement distress segmentation is accurately detecting and segmenting very small cracks or distresses. Identifying these tiny defects early enables preventive maintenance before they develop into more extensive damage. By intervening at this early stage, maintenance teams can prevent minor issues from escalating, thereby reducing repair costs and minimizing disruptions to traffic. To achieve these outcomes, it is essential to adopt advanced models capable of effectively handling both very small and larger cracks. However, achieving this goal is not without challenges, as many existing models struggle with multi-scale feature representation, which hinders their ability to effectively detect both small-scale and large-scale cracks\cite{Zhang2024-td}. In addition to this challenge is the lack of a comprehensive understanding of global context which often limits the model’s ability to capture large-scale spatial relationships and distinguish between interconnected distresses and noise. This results in inconsistent segmentation of extensive distress patterns such as longitudinal and alligator cracks. Furthermore, computational inefficiency compounds these issues, as processing the high-resolution images required for precise segmentation demands significant computational resources \cite{Huang2024-wy}. This challenge limits the feasibility of real-time deployment and scalability for large-scale pavement monitoring systems.  These above limitations highlight the pressing need for innovative approaches to enhance segmentation performance and address the shortcomings of existing methods.

To address the challenges of pavement distress segmentation, we propose Context-CrackNet, an encoder-decoder architecture built around two key innovations: the Region-Focused Enhancement Module (RFEM) and the Context-Aware Global Module (CAGM). The RFEM, embedded in the decoder pathway, prioritizes fine-grained features, enabling precise segmentation of small and tiny cracks. This ensures early detection of subtle distresses that traditional models often miss. The CAGM, positioned before the bottleneck, captures global context efficiently by integrating linear self-attention into its design. This allows the model to process high-resolution images and segment larger cracks, such as longitudinal and alligator cracks, without excessive computational costs.. The main contributions of our research has been outlined below:

\begin{itemize}
    \item Proposed \textbf{Context-CrackNet}, a novel efficient architecture that integrates specialized modules designed for comprehensive crack detection at varying scales in high-resolution pavement images. 

    \item Developed a \textbf{Region-Focused Enhancement Module} (RFEM) that employs targeted feature enhancement to capture fine-grained details of subtle cracks, enabling precise segmentation of small-scale pavement distress patterns.

    \item Introduced a \textbf{Context-Aware Global Module} (CAGM) that utilizes global contextual information to effectively identify and segment large-scale distress patterns while maintaining computational efficiency across high-resolution images.

    \item Trained and evaluated our proposed architecture on 10 publicly available crack datasets alongside several existing state-of-the-art segmentation models. Our proposed architecture consistently outperformed these models, achieving state-of-the-art performance across all benchmarks.
    
\end{itemize}

\section{Related Works}
\label{related-works}
Early attempts at pavement crack detection primarily relied on low-level image properties such as gradient, brightness, shape, and texture, as well as pixel intensity variance, edge orientation, and local binary patterns. Classic edge detection algorithms such as Sobel and Canny \cite{NHATDUC2018203},  Gabor filter-based methods \cite{gabor-filter, Zalama2013, Chen2020Gabor}, Prewitt \cite{Zhang2013Matched}, Roberts Cross \cite{Zhang2013Matched}, and Laplacian of Gaussian \cite{Dorafshan2019Benchmarking} identified crack characteristics by examining intensity variations and local directional patterns. Some researchers also employed threshold-based strategies such as Adaptive and Localized Thresholding \cite{Zhang2017An, Li2023Crack, 2022Using}, Triple-Thresholds Approach \cite{Peng2020A}, Otsu thresholding \cite{otsu, otsu-v2, otsu-v3} and wavelet transformations \cite{wavelet} to isolate cracks from background textures. Building on these foundations, early machine learning approaches  \cite{Liang2018An, Hoang2018A, Ahmadi2021An, Barkiah2023Overcoming, Müller2021Machine, ZOU2012227, Zhang2014AutomaticCD, random-forest-struct} framed crack extraction as a classification problem, distinguishing between crack and non-crack pixels using hand-engineered features.

However, these traditional methods often struggled with generalization \cite{Gao2020Autonomous}. Differences between training and testing datasets commonly led to performance degradation, and detecting tiny cracks proved particularly challenging. Moreover, the reliance on handcrafted features made these methods sensitive to environmental variations and morphological differences in cracks \cite {early-detection}. Noise and other forms of interference further undermined their stability and applicability.

With the rapid advancement of deep learning \cite{LeCun2015}, researchers turned to Convolutional Neural Networks (CNNs) \cite{cnns} to develop more robust solutions. CNNs excel at feature extraction, inspiring the creation of models for crack image classification, crack detection and crack segmentation. For instance, Li et al. \cite{crack-classification} proposed a CNN-based method to classify pavement patches into five categories using 3D pavement images, demonstrating high accuracy in distinguishing between the various crack types. Zhang et al. \cite{crack-first-seg} developed a deep convolutional neural network (ConvNet) for pavement crack detection, learning features directly from raw image patches and outperforming traditional hand-crafted approaches. Subsequently, Liu et al. \cite{deepcrack} proposed DeepCrack, a deep hierarchical CNN for pixel-wise crack segmentation, incorporating multi-scale feature fusion, deeply-supervised nets, and guided filtering learning methods that steadily improved segmentation performance.

Despite these advancements, significant challenges persisted. Many state-of-the-art methods emphasized performance metrics but gave limited attention to extracting subtle, tiny crack features. Additionally, as models became more complex, their computational and memory requirements increased, hindering deployment on resource-limited devices. Recognizing these issues, researchers began exploring lightweight model architectures capable of balancing detection accuracy with efficiency.

Li et al. \cite{carnet} proposed CarNet, a lightweight encoder-decoder achieving an ODS F-score of 0.514 on Sun520 with an inference speed of 104 FPS, balancing performance and efficiency. Similarly, Zhou et al. \cite{split-exchange} introduced LightCrackNet, a lightweight crack detection model designed to optimize performance and efficiency. The model utilizes Split Exchange Convolution (SEConv) and Multi-Scale Feature Exchange (MSFE) modules, achieving an F1-score of 0.867 DeepCrack dataset with only 1.3M parameters and 8 GFLOPs. In parallel, Omar et al. (Almaqtari et al. 2024) adopted a modular-based approach, including a Parallel Feature Module (PFM) and an Edge Extraction Module (EEM), to produce a model with just 0.87M parameters and 6.56G FLOPs, while attaining an F1-score of 0.864 and a comparable MIoU on the DeepCrack dataset.

Nevertheless, significant hurdles remain in achieving high-performance, efficient crack detection. Although CNN-based methods excel at extracting local features, they struggle to aggregate global context when used alone, limiting their ability to identify tiny or small cracks. Traditional self-attention transformer models offer a promising solution for capturing global relationships, but their quadratic scaling with input data makes them computationally intensive and impractical for certain applications. Linear self-attention transformers \cite{linformer, soft, Lee2018Set} when used with CNNs address this challenge by reducing computational complexity, enabling efficient integration of global and local cues while focusing on extracting minute crack features—a key objective in the field. Some researchers have applied linear self-attention modules in different domains. For instance, Fang et al. \cite{Fang2022CLFormer} proposed CLFormer, a lightweight transformer combining convolutional embedding and linear self-attention (LSA) for bearing fault diagnosis. It achieves strong robustness and high accuracy under noise and limited data, with only 4.88K parameters and 0.12 MFLOPs. Guo et al. \cite{Guo2021Beyond} introduced External Attention, a lightweight mechanism with linear complexity using two learnable memory layers, enhancing generalization and efficiency across visual tasks like segmentation and detection. 

\section{Method}
\subsection{Problem Structure and Overview}
Detecting small and tiny cracks in pavement images is particularly challenging due to their subtle features, irregular shapes, and the presence of noise such as shadows and debris. Existing deep learning models often struggle with capturing these fine-grained details, underscoring the need for more precise and efficient segmentation approaches. The problem definition has been formulated mathematically below. 

Let \( I \in \mathbb{R}^{H \times W \times C} \) represent a pavement image, where \( H \), \( W \), and \( C \) denote the height, width, and number of channels, respectively. The goal is to predict a segmentation map \( \hat{S} \in \mathbb{R}^{H \times W \times K} \), where \( K \) represents the number of classes, including the background. Mathematically, this can be expressed as learning a mapping function \( f: \mathbb{R}^{H \times W \times C} \to \mathbb{R}^{H \times W \times K} \), such that:
\begin{equation}
\hat{S} = f(I; \theta),
\end{equation}
where \( \theta \) denotes the learnable parameters of the segmentation model. The task involves accurately localizing and classifying pavement distresses of varying scales, shapes, and textures.

The proposed architecture, \textit{Context-CrackNet}, addresses these challenges by introducing two novel components: the Context-Aware Global Module (CAGM) and the Region-Focused Enhancement Module (RFEM). The CAGM ensures efficient global context modeling, enabling the network to capture long-range dependencies and large-scale spatial relationships. The RFEM enhances the network's ability to focus on fine-grained details, ensuring precise segmentation of small and subtle distresses. Together, these components form a robust encoder-decoder framework optimized for pavement distress segmentation.

\subsection{Overall Framework}

The overall structure of \textit{Context-CrackNet} integrates global and local feature refinement seamlessly, providing a balanced approach to segmentation. The network adopts an encoder-decoder structure, where the encoder extracts hierarchical features from the input image, the bottleneck incorporates global attention mechanisms, and the decoder reconstructs the segmentation map using refined skip connections.

\begin{figure}[h]
\centering
\includegraphics[width=1.00\textwidth]{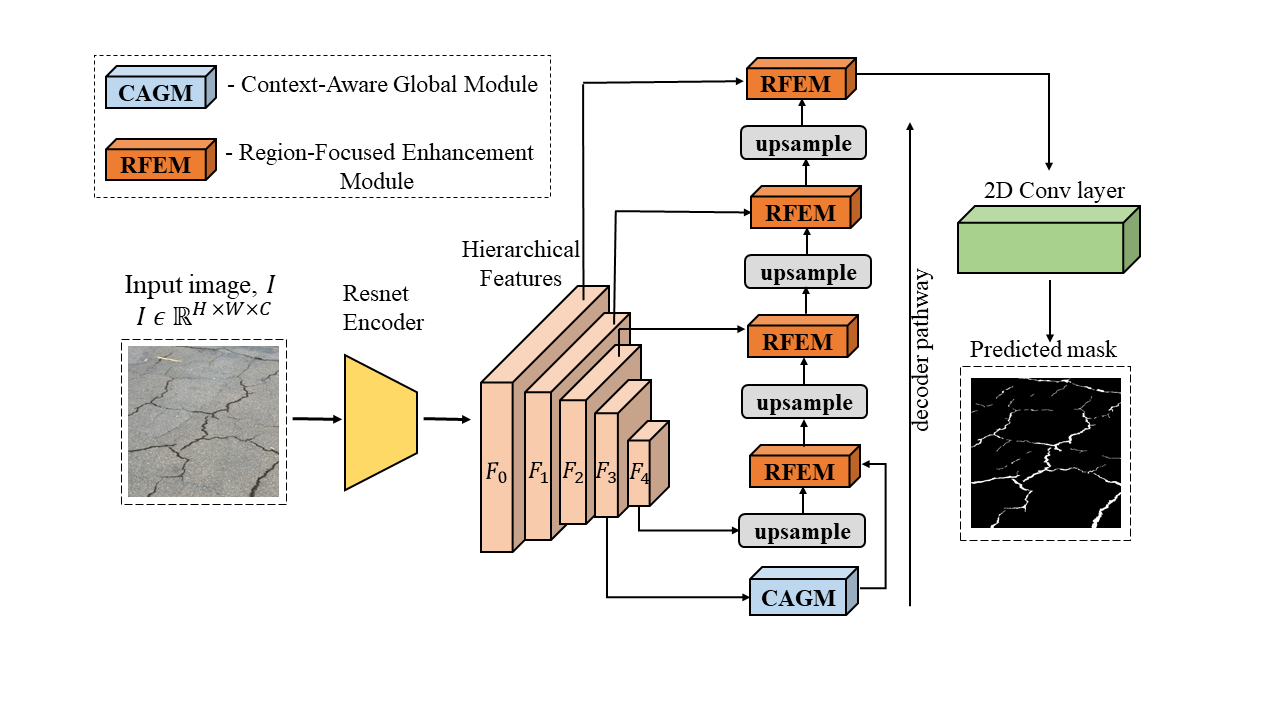}
\caption{Overall Architecture of \textit{Context-CrackNet}: The proposed framework adopts an encoder-decoder structure with two novel components: the Context-Aware Global Module (CAGM) and the Region-Focused Enhancement Module (RFEM). The ResNet-based encoder extracts hierarchical features \( \{F_0, F_1, F_2, F_3, F_4\} \), where \( F_3 \) is processed by the CAGM to model global contextual relationships and generate the contextualized feature map. The decoder pathway integrates RFEMs at each stage, which refine the skip connections between encoder features and upsampled decoder outputs. This refinement enables effective feature modulation for precise segmentation. Finally, the decoder outputs the predicted segmentation mask \( \hat{S} \), capturing fine-grained pavement distress details.}

\end{figure}

The encoder is based on a ResNet50 backbone, which extracts features at multiple levels of abstraction. For an input image \( I \), the encoder produces a sequence of feature maps:
\begin{equation}
\Phi_{\text{enc}}(I) = \{ F_0, F_1, F_2, F_3, F_4 \},
\end{equation}
where \( F_0 \in \mathbb{R}^{H/2 \times W/2 \times 64} \) represents low-level spatial features, and \( F_4 \in \mathbb{R}^{H/32 \times W/32 \times 2048} \) captures high-level semantic information. These feature maps progressively encode spatial and contextual details, forming the foundation for subsequent processing stages.

At the bottleneck, the feature map \( F_3 \), which encapsulates rich semantic information, is passed through the Context-Aware Global Module (CAGM). The CAGM uses a linear self-attention mechanism to model long-range dependencies efficiently, reducing the computational complexity typically associated with traditional self-attention. This produces an enhanced feature map \( F_{\text{CAGM}} \), expressed mathematically as:
\begin{equation}
F_{\text{CAGM}} = f_{\text{CAGM}}(F_3),
\end{equation}
where \( f_{\text{CAGM}} \) represents the operations within the module.

In the decoder pathway, the Region-Focused Enhancement Module (RFEM) plays a critical role in refining the skip connections between the encoder and decoder. For each decoder stage \( l \), the RFEM processes the corresponding encoder feature map \( F_{e,l} \) and the upsampled feature map \( F_{d,l+1} \) from the previous decoder stage. The refined feature map \( F_{\text{RFEM},l} \) is computed as:
\begin{equation}
F_{\text{RFEM},l} = f_{\text{RFEM}}(F_{e,l}, F_{d,l+1}),
\end{equation}
where \( f_{\text{RFEM}} \) represents the attention mechanism used to focus on the most relevant spatial regions. This refinement ensures that critical features are emphasized while irrelevant activations are suppressed.

The decoder reconstructs the segmentation map by iteratively combining the refined features from the RFEM with the upsampled feature maps. Starting with the output of the CAGM, the decoder applies a series of upsampling and refinement operations to produce the final segmentation map:
\begin{equation}
\hat{S} = f_{\text{decoder}}(F_{\text{RFEM}}),
\end{equation}
where \( f_{\text{decoder}} \) represents the decoding operations, including upsampling, concatenation, and convolution.

This framework effectively addresses the challenges associated with multi-scale feature representation and computational efficiency. By combining the strengths of the CAGM and RFEM, \textit{Context-CrackNet} achieves a balance between global context understanding and fine-grained detail enhancement, enabling robust and accurate pavement distress segmentation. The subsequent sections goes deeper into the mathematical details and implementation of the CAGM and RFEM modules.

\begin{algorithm}[t]
\caption{Context-CrackNet Framework}
\label{alg:Context-CrackNet}
\begin{algorithmic}[1]
\State Input image $I \in \mathbb{R}^{H \times W \times C}$
\State Predicted segmentation mask $\hat{S} \in \mathbb{R}^{H \times W \times K}$
\State \textbf{Stage 1: Encoder Pathway}
\State $\{F_0, F_1, F_2, F_3, F_4\} \leftarrow \text{ResNetEncoder}(I)$ \Comment{Extract features}
\State \textbf{Stage 2: Bottleneck Processing}
\State $F_{\text{CAGM}} \leftarrow \text{CAGM}(F_3)$ \Comment{Global context modeling}
\State \textbf{Stage 3: Decoder Pathway}
\State Initialize $D_4 \leftarrow F_{\text{CAGM}}$ $l \in \{3, 2, 1, 0\}$
\State $D_{\text{up}} \leftarrow \text{Upsample}(D_{l+1})$ \Comment{2$\times$ spatial size}
\State $D_l \leftarrow \text{RFEM}(F_l, D_{\text{up}})$ \Comment{Feature refinement}
\State \textbf{Stage 4: Final Prediction}
\State $\hat{S} \leftarrow \text{Conv2D}(D_0)$ \Comment{K-class prediction}
\State $\hat{S}$
\end{algorithmic}
\end{algorithm}

\subsection{Context-Aware Global Module (CAGM)}

The \textbf{Context-Aware Global Module (CAGM)} addresses the challenge of capturing long-range dependencies and global contextual relationships in the feature map \( F_3 \). This capability is crucial for accurately segmenting large-scale pavement distresses, such as longitudinal and alligator cracks, which require an understanding of spatial relationships across distant regions. To achieve this, the CAGM employs a linear self-attention mechanism, reducing the quadratic complexity of traditional self-attention to linear, thereby enabling efficient processing of high-resolution images.

\begin{figure}[h]
\centering
\includegraphics[width=0.8\textwidth]{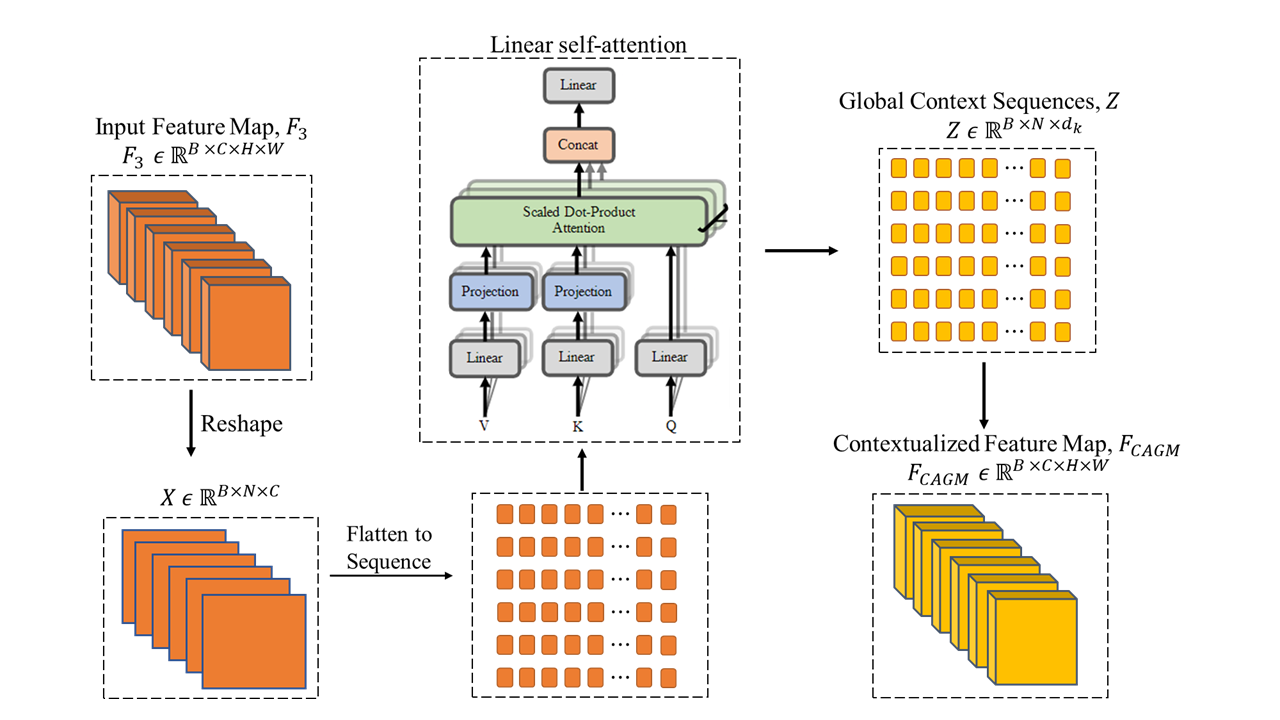}
\caption{Context-Aware Global Module (CAGM): The module processes the input feature map \( F_3 \in \mathbb{R}^{B \times C \times H \times W} \), reshaping it into a sequence \( X \in \mathbb{R}^{B \times N \times C} \), where \( N = H \times W \). Using a \textbf{Linear Self-Attention Mechanism}, query (\( Q \)), key (\( K \)), and value (\( V \)) projections generate \textbf{Global Context Sequences} (\( Z \in \mathbb{R}^{B \times N \times d_k} \)). These sequences are reconstructed into the \textbf{Contextualized Feature Map} (\( F_{\text{CAGM}} \in \mathbb{R}^{B \times C \times H \times W} \)), embedding global dependencies efficiently.}

\end{figure}

Let \( F_3 \in \mathbb{R}^{B \times C \times H \times W} \) represent the input feature map at the bottleneck stage, where \( B \) is the batch size, \( C \) is the number of channels, and \( H, W \) are the spatial dimensions. The first step in the CAGM is to transform the spatial dimensions \( H \) and \( W \) into a sequence of length \( N = H \times W \). This can be mathematically expressed as:
\begin{equation}
X_b = \{ F_3[b, :, i, j] \mid i \in \{1, \ldots, H\}, j \in \{1, \ldots, W\} \}, \quad X \in \mathbb{R}^{B \times N \times C},
\end{equation}
where \( X_b \) is the reshaped feature map for the \( b \)-th sample in the batch, and \( X \) concatenates all spatial positions into a sequence.

The sequence \( X \) is projected into query (\( Q \)), key (\( K \)), and value (\( V \)) spaces using learned linear transformations:
\begin{equation}
Q = X W_Q, \quad K = X W_K, \quad V = X W_V,
\end{equation}
where \( W_Q, W_K, W_V \in \mathbb{R}^{C \times d_k} \) are learnable weight matrices, and \( d_k \) is the dimensionality of the query and key vectors.

To reduce the computational cost, the key and value matrices are projected into lower-dimensional spaces:
\begin{equation}
K_{\text{proj}} = K E, \quad V_{\text{proj}} = V F,
\end{equation}
where \( E, F \in \mathbb{R}^{N \times k} \) are learnable projection matrices, and \( k \ll N \) is the reduced dimension of the projected key and value spaces.

The attention weights are computed as:
\begin{equation}
A = \text{softmax}\left(\frac{Q K_{\text{proj}}^\top}{\sqrt{d_k}}\right), \quad A \in \mathbb{R}^{B \times N \times k},
\end{equation}
where \( \text{softmax} \) ensures the weights \( A \) sum to 1 across the key dimension for each query.

The weighted output is then computed by aggregating the projected values:
\begin{equation}
Z = A V_{\text{proj}}, \quad Z \in \mathbb{R}^{B \times N \times d_k}.
\end{equation}

Finally, the output sequence \( Z \) is mapped back to the original channel dimension and reconstructed into its spatial structure:
\begin{equation}
F_{\text{CAGM}}[b, :, i, j] = Z[b, n] W_O, \quad n = (i-1) \times W + j,
\end{equation}
where \( W_O \in \mathbb{R}^{d_k \times C} \) is a learnable weight matrix, and \( F_{\text{CAGM}} \in \mathbb{R}^{B \times C \times H \times W} \) is the enhanced feature map.

By explicitly modeling global relationships across spatial regions, the CAGM integrates information from distant parts of the pavement image, enabling the network to detect and segment large-scale cracks and patterns. Its efficient linear self-attention mechanism ensures scalability, making it suitable for high-resolution images while maintaining computational feasibility.



\begin{algorithm}[t]
\caption{Context-Aware Global Module (CAGM)}
\label{alg:cagm}
\begin{algorithmic}[1]
\Require
    \State Input feature map $F_3 \in \mathbb{R}^{B \times C \times H \times W}$
\Ensure
    \State Contextualized feature map $F_{\text{CAGM}} \in \mathbb{R}^{B \times C \times H \times W}$

\vspace{1mm}
\State \textbf{Feature Reshaping:}
    \State $X \leftarrow \text{Reshape}(F_3)$ \Comment{$X \in \mathbb{R}^{B \times N \times C}$, where $N = H \times W$}

\vspace{1mm}
\State \textbf{Linear Self-Attention:}
    \State $Q \leftarrow \text{Linear}(X)$ \Comment{Query projection}
    \State $K \leftarrow \text{Linear}(X)$ \Comment{Key projection}
    \State $V \leftarrow \text{Linear}(X)$ \Comment{Value projection}
    \State $A \leftarrow \text{SDP-Attention}(Q, K, V)$ \Comment{Scaled dot-product attention}
    \State $Z \leftarrow \text{Concat}[A, X]$ \Comment{$Z \in \mathbb{R}^{B \times N \times d_k}$}
    \State $Z \leftarrow \text{Linear}(Z)$ \Comment{Final projection}

\vspace{1mm}
\State \textbf{Global Context Reconstruction:}
    \State $F_{\text{CAGM}} \leftarrow \text{Reshape}(Z, [B, C, H, W])$ \Comment{Restore spatial dimensions}

\Return $F_{\text{CAGM}}$

\Statex \textbf{Note:} SDP-Attention computes $\text{softmax}(\frac{QK^T}{\sqrt{d_k}})V$
\end{algorithmic}
\end{algorithm}

\subsection{Region-Focused Enhancement Module (RFEM)}

The \textbf{Region-Focused Enhancement Module (RFEM)} refines the skip connections between the encoder and decoder to enhance the segmentation of fine-grained details such as small and subtle pavement cracks. By dynamically modulating encoder features using spatial context from the decoder, the RFEM ensures that relevant features are emphasized, while irrelevant ones are suppressed. This capability is critical for capturing the intricate structures of small distresses that traditional methods often overlook.

\begin{figure}[h]
\centering
\includegraphics[width=0.9\textwidth]{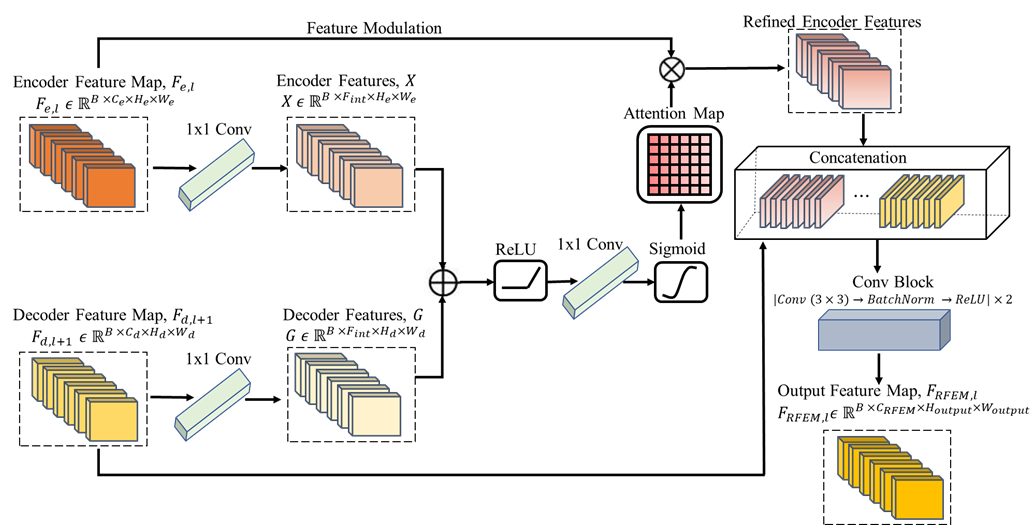}
\caption{Region-Focused Enhancement Module (RFEM): The module refines encoder features \( F_{e,l} \in \mathbb{R}^{B \times C_e \times H_e \times W_e} \) and decoder features \( F_{d,l+1} \in \mathbb{R}^{B \times C_d \times H_d \times W_d} \) by transforming them into intermediate features \( X \) and \( G \), respectively, via \( 1 \times 1 \) convolutions. These are combined to compute an \textbf{Attention Map} through element-wise addition, a ReLU activation, and a sigmoid activation. The attention-modulated encoder features are concatenated with the upsampled decoder features and passed through a \textbf{Conv Block} (\( \text{Conv} (3 \times 3) \to \text{BatchNorm} \to \text{ReLU} \) repeated twice), producing the refined output feature map \( F_{\text{RFEM},l} \in \mathbb{R}^{B \times C_{\text{RFEM}} \times H_{\text{output}} \times W_{\text{output}}} \). This process emphasizes fine-grained details while maintaining contextual relevance.}

\end{figure}

Let \( F_{e,l} \in \mathbb{R}^{B \times C_e \times H_e \times W_e} \) represent the feature map from the encoder at stage \( l \), and \( F_{d,l+1} \in \mathbb{R}^{B \times C_d \times H_d \times W_d} \) represent the upsampled feature map from the preceding decoder stage. The RFEM computes a refined feature map \( F_{\text{RFEM},l} \) by integrating attention-driven feature enhancement and concatenation as follows:

First, the encoder and decoder feature maps are linearly transformed into a shared intermediate space:
\begin{equation}
G = F_{d,l+1} \ast W_g + b_g, \quad X = F_{e,l} \ast W_x + b_x,
\end{equation}
where \( W_g, W_x \) are learnable convolutional kernels, and \( b_g, b_x \) are corresponding biases. These transformations ensure compatibility between the encoder and decoder feature maps in terms of dimensionality and spatial resolution.

The transformed features are then combined via element-wise addition, followed by the application of a non-linear activation function to generate an attention activation map:
\begin{equation}
Y = \max(0, G + X),
\end{equation}
where the \(\max(0, \cdot)\) operation ensures non-linearity by setting all negative values to zero. This introduces the capability to model complex spatial interactions effectively.

To focus on the most salient regions, the module computes an attention coefficient map \( \Psi \) through a linear transformation followed by a sigmoid activation:
\begin{equation}
\Psi = \frac{1}{1 + \exp\left(-\left(Y \ast W_\psi + b_\psi\right)\right)},
\end{equation}
where \( W_\psi \) and \( b_\psi \) are learnable parameters. The sigmoid function normalizes the attention coefficients to the range \([0, 1]\), representing the relative importance of each spatial location.

The encoder feature map is refined by modulating it with the attention coefficients:
\begin{equation}
F_{\text{refined}} = \Psi \odot F_{e,l},
\end{equation}
where \( \odot \) denotes element-wise multiplication, ensuring that only the most critical features are retained for subsequent processing.

The refined encoder feature map is combined with the upsampled decoder feature map along the channel dimension:
\begin{equation}
F_{\text{concat}}[b, c, i, j] =
\begin{cases} 
F_{\text{refined}}[b, c, i, j], & \text{for } c < C_{\text{refined}}, \\
F_{d,l+1}[b, c - C_{\text{refined}}, i, j], & \text{for } c \geq C_{\text{refined}},
\end{cases}
\end{equation}
where \( C_{\text{refined}} \) is the number of channels in \( F_{\text{refined}} \). This operation combines the spatially refined encoder features with the contextual information from the decoder by aligning them along the channel axis.

Finally, the concatenated features are processed through a convolutional block to produce the updated decoder feature map:
\begin{equation}
F_{\text{RFEM},l} = \gamma(F_{\text{concat}}),
\end{equation}
where \( \gamma(\cdot) \) represents the operations of the convolutional block, typically including convolution, normalization, and activation functions.

By focusing on spatially relevant features and suppressing noise, the RFEM significantly enhances the model's ability to detect and segment fine-grained pavement distresses, such as hairline cracks, that are often missed by conventional approaches. This module complements the global context modeling of the CAGM by ensuring that local, fine-scale details are not overshadowed by broader spatial relationships.



\begin{algorithm}[t]
\caption{Region-Focused Enhancement Module (RFEM)}
\label{alg:rfem}
\begin{algorithmic}[1]
\Require
    \State Encoder feature map $F_{e,l} \in \mathbb{R}^{B \times C_e \times H_e \times W_e}$
    \State Decoder feature map $F_{d,l+1} \in \mathbb{R}^{B \times C_d \times H_d \times W_d}$
\Ensure
    \State Refined feature map $F_{\text{RFEM},l} \in \mathbb{R}^{B \times C_{\text{RFEM}} \times H_{\text{output}} \times W_{\text{output}}}$

\vspace{1mm}
\State \textbf{Feature Transformation:}
    \State $X \leftarrow \text{Conv}_{1\times1}(F_{e,l})$ \Comment{Transform to $\mathbb{R}^{B \times F_{\text{int}} \times H_e \times W_e}$}
    \State $G \leftarrow \text{Conv}_{1\times1}(F_{d,l+1})$ \Comment{Transform to $\mathbb{R}^{B \times F_{\text{int}} \times H_d \times W_d}$}

\vspace{1mm}
\State \textbf{Attention Map Generation:}
    \State $Y \leftarrow \text{ReLU}(G + X)$ \Comment{Element-wise addition}
    \State $\Psi \leftarrow \sigma(\text{Conv}_{1\times1}(Y))$ \Comment{$\sigma$: Sigmoid activation}

\vspace{1mm}
\State \textbf{Feature Modulation:}
    \State $F_{\text{refined}} \leftarrow \Psi \otimes F_{e,l}$ \Comment{Channel-wise multiplication}

\vspace{1mm}
\State \textbf{Feature Fusion:}
    \State $F_{\text{concat}} \leftarrow \text{Concat}([F_{\text{refined}}, F_{d,l+1}])$

\vspace{1mm}
\State \textbf{Conv Block Refinement:}
    \For{$i \leftarrow 1$ \textbf{to} $2$}
        \State $F_{\text{RFEM},l} \leftarrow \text{ReLU}(\text{BatchNorm}(\text{Conv}_{3\times3}(F_{\text{concat}})))$
    \EndFor

\Return $F_{\text{RFEM},l}$
\end{algorithmic}
\end{algorithm}

\subsection{Loss Functions}

The proposed framework uses tailored loss functions to optimize the segmentation task across both binary and multi-class scenarios, ensuring accurate prediction of pavement distress patterns. These loss functions are carefully designed to balance class contributions, address class imbalance, and effectively capture fine-grained details. 

\subsubsection{Binary Segmentation Loss}
For binary segmentation tasks, where the goal is to classify each pixel as either belonging to a crack (\(1\)) or not (\(0\)), we employ a combination of the Binary Cross Entropy (BCE) loss and the Dice loss. The combined loss is formulated as:
\begin{equation}
\mathcal{L}_{\text{binary}} = \alpha \, \mathcal{L}_{\text{BCE}} + \beta \, \mathcal{L}_{\text{Dice}},
\end{equation}
where \( \alpha \) and \( \beta \) are weights that control the contribution of each term.

\paragraph{1. Binary Dice Loss}
The Dice loss, designed to handle imbalances in pixel classes, measures the overlap between the predicted segmentation map \( \hat{S} \) and the ground truth \( S \). 

Let \( \hat{S} \in [0, 1]^{H \times W} \) and \( S \in \{0, 1\}^{H \times W} \) denote the predicted and ground truth maps, respectively. The Dice loss is computed as:
\begin{equation}
\mathcal{L}_{\text{Dice}} = 1 - \frac{2 \, \sum_{i=1}^{N} \hat{S}_i S_i + \epsilon}{\sum_{i=1}^{N} \hat{S}_i + \sum_{i=1}^{N} S_i + \epsilon},
\end{equation}
where \( \epsilon \) is a smoothing constant to prevent division by zero, and \( N = H \times W \) represents the total number of pixels. The Dice loss encourages the model to maximize the overlap between \( \hat{S} \) and \( S \), ensuring robust segmentation of small and subtle cracks.

\paragraph{2. Binary Cross Entropy Loss}
The BCE loss penalizes the deviation between predicted probabilities and ground truth labels. It is defined as:
\begin{equation}
\mathcal{L}_{\text{BCE}} = - \frac{1}{N} \sum_{i=1}^{N} \left[ S_i \log(\hat{S}_i) + (1 - S_i) \log(1 - \hat{S}_i) \right].
\end{equation}
This term provides pixel-wise supervision, complementing the Dice loss by ensuring accurate classification even in cases of severe class imbalance.

\subsubsection{Multi-Class Segmentation Loss}
For multi-class segmentation tasks, where the pavement image contains multiple types of distresses, we adopt a combined loss function comprising the Cross-Entropy (CE) loss and a multi-class Dice loss:
\begin{equation}
\mathcal{L}_{\text{multi-class}} = \gamma \, \mathcal{L}_{\text{CE}} + \delta \, \mathcal{L}_{\text{Dice}},
\end{equation}
where \( \gamma \) and \( \delta \) are weighting factors.

\paragraph{1. Multi-Class Dice Loss}  
The Multi-Class Dice Loss extends the principles of the Binary Dice Loss to multi-class segmentation tasks, ensuring fair optimization for each class, including the background. By individually evaluating the overlap between the predicted segmentation map \( \hat{S}_k \) and the ground truth map \( S_k \) for each class \( k \), it addresses class imbalances and promotes accurate segmentation across all categories.

Let \( \hat{S}_k \) and \( S_k \) represent the predicted and ground truth maps for class \( k \), respectively, where \( k \in \{1, \dots, K\} \). The Dice loss for class \( k \) is defined as:
\begin{equation}
\mathcal{L}_{\text{Dice},k} = 1 - \frac{2 \, \sum_{i=1}^{N} \hat{S}_{k,i} S_{k,i} + \epsilon}{\sum_{i=1}^{N} \hat{S}_{k,i} + \sum_{i=1}^{N} S_{k,i} + \epsilon},
\end{equation}
where \( N \) denotes the total number of pixels and \( \epsilon \) is a small constant to avoid division by zero.

The total multi-class Dice loss is calculated as the average Dice loss across all \( K \) classes:
\begin{equation}
\mathcal{L}_{\text{Dice}} = \frac{1}{K} \sum_{k=1}^{K} \mathcal{L}_{\text{Dice},k}.
\end{equation}
This formulation encourages precise segmentation across all classes, ensuring small or underrepresented categories are effectively captured.

\paragraph{2. Cross-Entropy Loss}
The CE loss measures the pixel-wise classification error, weighted by class frequencies to handle imbalance. 

Let \( \omega_k \) denote the weight for class \( k \), derived from the inverse class frequency. The CE loss is defined as:
\begin{equation}
\mathcal{L}_{\text{CE}} = - \frac{1}{N} \sum_{i=1}^{N} \sum_{k=1}^{K} \omega_k S_{k,i} \log(\hat{S}_{k,i}),
\end{equation}
where \( \omega_k \) ensures that the model does not bias toward dominant classes.

\section{Experiments}
To validate the effectiveness of the proposed \textit{Context-CrackNet}, we conduct comprehensive experiments designed to evaluate its performance on pavement distress segmentation tasks. This section details the datasets used, the implementation specifics of \textit{Context-CrackNet}, and the experimental setup. Furthermore, ablation studies are performed to assess the contribution of each module and design choice to the overall performance. Finally, we compare \textit{Context-CrackNet} with state-of-the-art methods to highlight its advantages in addressing the challenges of fine-grained and multi-scale pavement distress detection.

\subsection{Dataset}
\label{sec:dataset}
To train and evaluate the proposed \textit{Context-CrackNet}, we utilize 10 publicly available binary crack datasets: CFD \cite{CFD}, Crack500 \cite{CRACK500}, CrackTree200 \cite{cracktree200}, DeepCrack \cite{deepcrack}, Eugen Miller \cite{eugen_miller}, Forest \cite{forest}, GAPs \cite{GAPS384}, Rissbilder \cite{Volker}, Sylvie \cite{sylvie}, and Volker \cite{Volker}. These datasets comprehensively cover diverse crack detection scenarios, including road cracks, concrete cracks in tunnels, and wall cracks. This diversity demonstrates the robustness of \textit{Context-CrackNet} beyond pavement applications, highlighting its potential to handle various types of cracks across different construction materials and structures.

\begin{table}[htbp]
    \centering
    \caption{Datasets and their corresponding crack types}
    \label{tab:crack_types}
    \begin{tabular}{ll}
    \hline
    \textbf{Dataset name} & \textbf{Crack type} \\
    \hline
    CFD & Road crack \\
    CRACK500 & Road crack \\
    CrackTree200 & Road crack \\
    DeepCrack & Road crack \\
    Forest & Road crack \\
    GAPs & Road crack \\
    Sylvie & Road crack \\
    Rissbilder & Wall crack \\
    Volker & Wall crack \\
    Eugen Miller & Concrete crack on Tunnels \\
    \hline
    \end{tabular}
\end{table}

The datasets exhibit a significant class imbalance, with approximately 97.2\% of pixels belonging to the background class and only 2.8\% corresponding to the crack class. This imbalance reflects the real-world challenges of identifying subtle cracks against vast background regions.

For uniformity, all images and corresponding masks were resized to a resolution of \( 448 \times 448 \) pixels. The datasets were then split into training and testing sets with an 80:20 ratio, ensuring a balanced distribution of samples across both sets.

To further enhance the diversity of the training data and improve the model's generalization capabilities, a series of data augmentation techniques were employed. These augmentations included spatial transformations such as horizontal and vertical flips, random rotations by \( 90^\circ \), and shift-scale-rotate operations, which varied the spatial properties of the images while maintaining their structural integrity. Additionally, pixel-level augmentations such as Gaussian noise and color jittering were applied to introduce variations in brightness, contrast, saturation, and hue, simulating real-world variations in lighting and camera conditions.

\begin{figure}[H]
\centering
\includegraphics[width=0.9\textwidth]{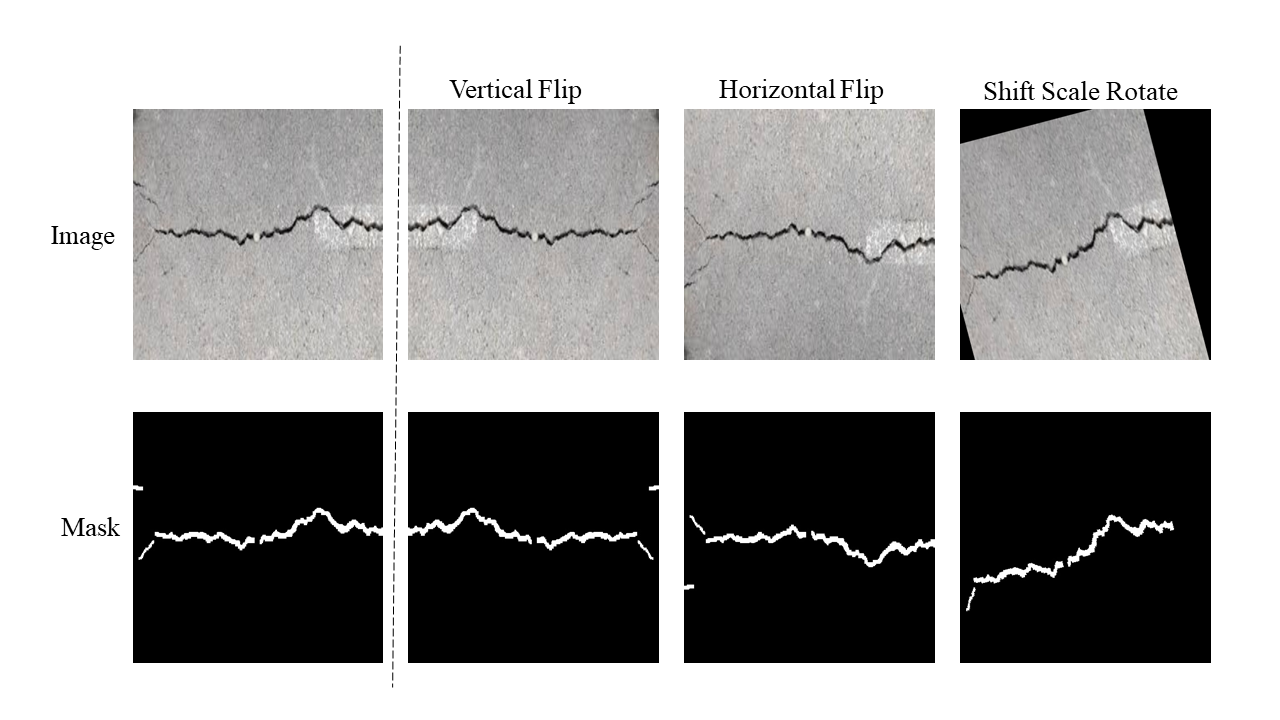}
\caption{Examples of data augmentation techniques applied to crack images and their corresponding masks in the DeepCrack dataset. Augmentations include vertical flip, horizontal flip, and shift-scale-rotate, showcasing the spatial transformations employed to enhance diversity and robustness in the training dataset. The top row illustrates augmented images, while the bottom row presents their respective masks.}
\end{figure}

For preprocessing, the images were normalized using the mean and standard deviation values of the ResNet backbone: \( \mu (0.485, 0.456, 0.406) \) and \( \sigma (0.229, 0.224, 0.225) \), respectively. This normalization step ensures compatibility with the pre-trained ResNet model used in the encoder.

By employing a diverse dataset that encompasses road, wall, and concrete cracks, the proposed framework is equipped to handle the complexities of real-world crack detection scenarios, addressing challenges such as class imbalance, varying scales, and noise effectively.

\subsubsection{Custom Dataset}

\subsection{Implementation details}
\subsubsection{Training settings}
The proposed \textit{Context-CrackNet} was implemented using the PyTorch framework. The AdamW \cite{AdamW} optimizer was used for training, with a weight decay of \( 1 \times 10^{-5} \)
 to prevent overfitting. The initial learning rate was set to \( 1 \times 10^{-4} \)
, and a  using an adaptive learning rate scheduler was applied to adjust the learning rate dynamically. This scheduler reduced the learning rate by a factor of 0.5 after 5 epochs of no improvement in validation loss. The model was then trained with a batch size of 32 for a total of 1000 epochs. All experiments were conducted on an NVIDIA A40 GPU with 48GB of memory, which provided the computational capacity to handle high-resolution images efficiently. Table 1 shows all the different training configurations used. 

\subsubsection{Evaluation metrics}

To comprehensively evaluate the performance of the proposed \textit{Context-CrackNet} on pavement distress segmentation tasks, we employed the following segmentation metrics: Intersection over Union (IoU) score, Dice score, Precision, Recall, and F1 score. These metrics provide a holistic assessment of the model’s ability to accurately segment fine-grained and multi-scale pavement cracks, balancing considerations of overlap, correctness, and completeness.

\paragraph{\textbf{Mean Intersection over Union (mIoU)}}
The mean Intersection over Union (mIoU) evaluates the average overlap between the predicted segmentation map \( \hat{S}_k \) and the ground truth \( S_k \) across all classes \( k \). For a single class \( k \), the IoU is defined as:
\[
\text{IoU}_k = \frac{| \hat{S}_k \cap S_k |}{| \hat{S}_k \cup S_k |} = \frac{\sum_{i=1}^N \hat{S}_{k,i} S_{k,i}}{\sum_{i=1}^N (\hat{S}_{k,i} + S_{k,i} - \hat{S}_{k,i} S_{k,i})},
\]
where \( N \) denotes the total number of pixels, and \( \hat{S}_{k,i}, S_{k,i} \in \{0, 1\} \) represent the predicted and ground truth labels for pixel \( i \) in class \( k \). The mIoU is then computed as the average IoU across all \( K \) classes:
\[
\text{mIoU} = \frac{1}{K} \sum_{k=1}^K \text{IoU}_k.
\]
This metric provides a comprehensive assessment of segmentation performance by considering the overlap for all classes and averaging them to yield a single performance score.

\paragraph{\textbf{Dice Score}}
The Dice score, also known as the Sørensen–Dice coefficient, quantifies the overlap between the predicted and ground truth segmentation maps. It is defined as:
\begin{equation}
\text{Dice} = \frac{2 | \hat{S} \cap S |}{| \hat{S} | + | S |} = \frac{2 \sum_{i=1}^N \hat{S}_i S_i}{\sum_{i=1}^N \hat{S}_i + \sum_{i=1}^N S_i}.
\end{equation}
Dice score emphasizes the correct segmentation of smaller regions, making it particularly useful for evaluating fine-grained crack details.

\paragraph{\textbf{Precision}}
Precision measures the proportion of correctly identified crack pixels to the total predicted crack pixels. It is defined as:
\begin{equation}
\text{Precision} = \frac{\text{True Positives (TP)}}{\text{True Positives (TP)} + \text{False Positives (FP)}}.
\end{equation}

\paragraph{\textbf{Recall}}
Recall quantifies the ability of the model to detect all crack pixels in the ground truth. It is defined as:
\begin{equation}
\text{Recall} = \frac{\text{True Positives (TP)}}{\text{True Positives (TP)} + \text{False Negatives (FN)}}.
\end{equation}

\paragraph{\textbf{F1 Score}}
The F1 score provides a harmonic mean of Precision and Recall, balancing their trade-offs. It is expressed as:
\begin{equation}
\text{F1} = 2 \cdot \frac{\text{Precision} \cdot \text{Recall}}{\text{Precision} + \text{Recall}}.
\end{equation}

These metrics collectively evaluate the model’s segmentation performance, ensuring both spatial accuracy (IoU, Dice) and the balance between prediction correctness and completeness (Precision, Recall, F1).

\begin{table}[h!]
\centering
\caption{Model training settings.}
\begin{tabular}{ll}
\hline
\textbf{Name} & \textbf{Training setting} \\ \hline
Optimizer & AdamW \\
Learning Rate & \( 1 \times 10^{-4} \) \\
Batch Size & 32 \\
Weight Decay & \( 1 \times 10^{-5} \) \\
Number of Epochs & 1000 \\
\hline
\end{tabular}
\label{tab:training_settings}
\end{table}

\subsubsection{Comparison with other methods}

To evaluate the performance of the proposed \textit{Context-CrackNet}, we compared it against state-of-the-art segmentation models, including U-Net \cite{Unet}, U-Net++ \cite{Unet++}, DeepLabV3 \cite{DeepLabV3}, DeepLabV3+ \cite{deeplabv3plus}, FPN \cite{FPN}, PSPNet \cite{PSPNet}, LinkNet \cite{Linknet}, MAnet \cite{MAnet}, and PAN \cite{PAN}. All models were trained and evaluated on the same 10 crack detection datasets (see Section~\ref{sec:dataset}) under identical experimental conditions to ensure fairness. Each method used a ResNet50 encoder pre-trained on ImageNet, consistent with \textit{Context-CrackNet}.

The datasets presented diverse challenges such as varying crack patterns, scales, and noise levels, providing a robust basis for comparison. Standardized training settings, including preprocessing, augmentations, and hyperparameters, ensured that performance differences reflected the strengths of the architectures rather than experimental inconsistencies.

\section{Results and Discussion}
This section presents the results of the proposed \textit{Context-CrackNet} across multiple datasets. Both qualitative and quantitative evaluations are discussed, highlighting the model's performance in comparison with existing state-of-the-art methods.

\subsection{Qualitative Analysis of Predictions}
In this section, we analyze the qualitative results of \textit{Context-CrackNet} compared to existing models, including MAnet, PSPNet, DeepLabV3+, and FPN. These results are evaluated across the ten diverse datasets containing different types of cracks, such as road cracks, wall cracks, and concrete cracks. The goal is to assess each model’s ability to detect both prominent and tiny cracks, which are critical for reliable structural assessment.

\begin{figure}[H]
\centering
\includegraphics[width=0.99\textwidth]{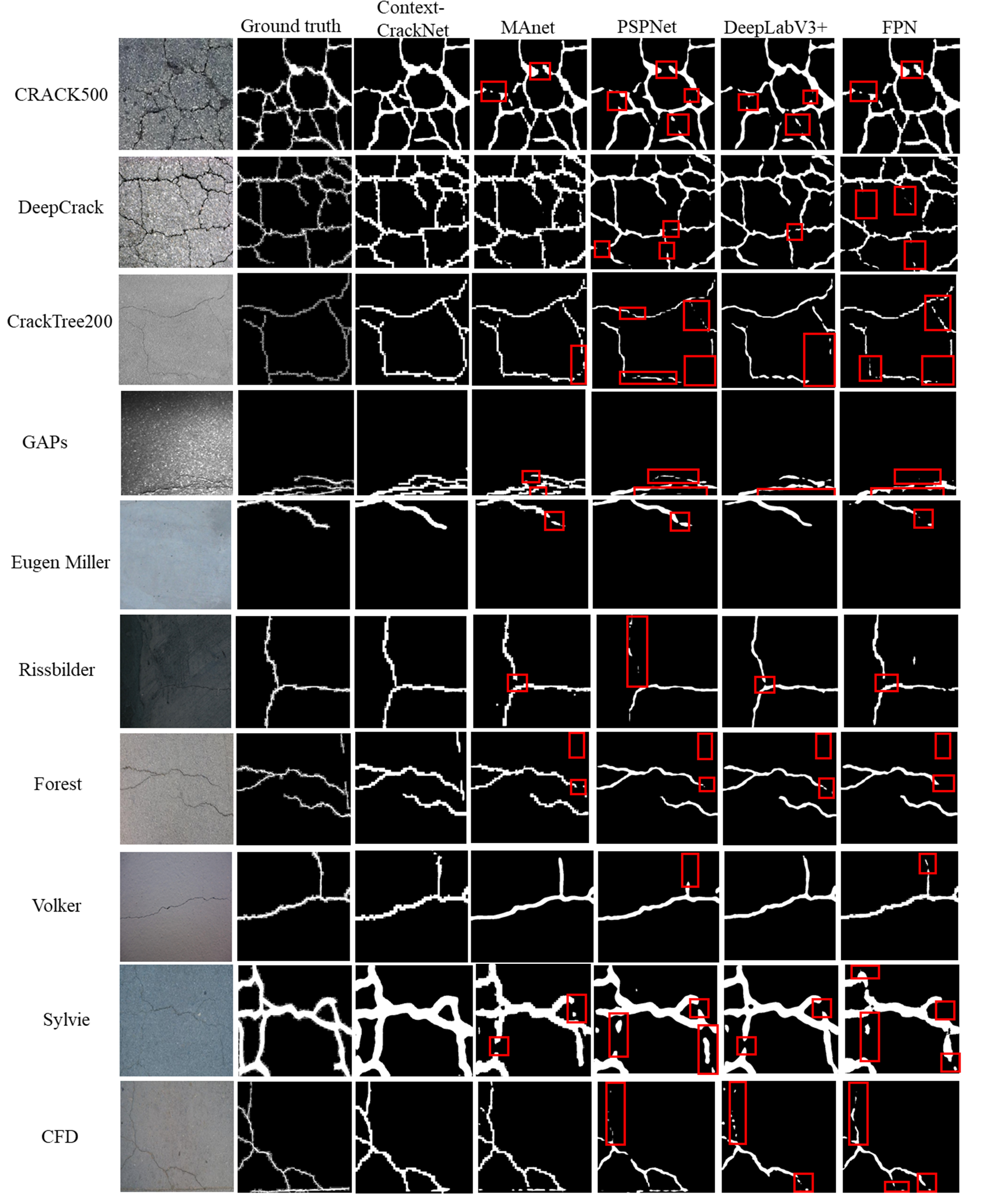}
\caption{Visual comparison of crack detection results across various datasets. The first column displays the input images, followed by the ground truth masks in the second column. The third column shows the predictions of the proposed \textit{Context-CrackNet}, while subsequent columns present predictions from comparison models including MAnet, PSPNet, DeepLabV3+, and FPN. Rows correspond to individual datasets (e.g., CRACK500, DeepCrack, CrackTree200, etc.). The red boxes highlight areas where comparison models fail to detect tiny cracks effectively, demonstrating the superior performance of \textit{Context-CrackNet} in accurately capturing fine-grained crack details.}
\label{fig:comparison_results}
\end{figure}

Figure~\ref{fig:comparison_results} shows segmentation results from various datasets. The first column displays the original crack images, followed by their ground truth masks. The subsequent columns show the predictions from \textit{Context-CrackNet} and other models. Each row represents a dataset, showcasing the models’ performance across different types of cracks.

The predictions demonstrate that \textit{Context-CrackNet} performs consistently better in detecting fine, small, and subtle cracks compared to the other models. For instance, in the CRACK500 dataset, \textit{Context-CrackNet} successfully identifies the small, interconnected cracks, which competing models often fail to detect (highlighted by red boxes). A similar trend is observed in the Rissbilder dataset, where \textit{Context-CrackNet} captures the thin wall cracks more accurately, while other models struggle with false positives or incomplete predictions.

\subsubsection{Performance on Diverse Crack Types}

The results emphasize \textit{Context-CrackNet}'s adaptability to different crack types and contexts. In the DeepCrack dataset, characterized by dense and complex crack patterns, \textit{Context-CrackNet} captures the overall structure of the cracks more effectively than models like PSPNet and DeepLabV3+, which tend to miss faint connections. Likewise, in the Eugen Miller dataset featuring tunnel cracks, \textit{Context-CrackNet} produces cleaner and more detailed predictions, showing its robustness on surfaces with uniform textures.

\subsubsection{Tiny Crack Detection and Generalization}

A key strength of \textit{Context-CrackNet} is its ability to detect tiny cracks that are often missed by other models. The red-marked areas in the predictions from competing models highlight their failure to consistently capture these smaller cracks. This reinforces the effectiveness of \textit{Context-CrackNet}'s context-aware design, which allows it to focus on both small-scale details and larger crack patterns. This capability addresses the central challenge of detecting tiny cracks, which are crucial for identifying early signs of structural damage.



\subsection{Quantitative Results}
The quantitative results in Tables \ref{tab:combined_results}, \ref{tab:combined_results_2}, \ref{tab:combined_results_3} demonstrate that \textit{Context-CrackNet} effectively addresses the challenge of detecting tiny and complex cracks, often missed by existing models. On the CFD dataset, it achieves the highest mIoU (\textbf{0.5668}) and Dice Score (\textbf{0.7235}), along with a recall of \textbf{0.8989}, showcasing its ability to capture subtle crack details where models like DeepLabV3 and FPN fall short.

In the CRACK500 dataset, which features diverse crack patterns, \textit{Context-CrackNet} outperforms others with an mIoU of \textbf{0.6733} and Dice Score of \textbf{0.8046}, demonstrating robust generalization to varying pavement conditions. Similarly, on CrackTree200, characterized by sparse and narrow cracks, it achieves the highest recall (\textbf{0.9386}) and Dice Score (\textbf{0.6992}), proving its sensitivity to fine-grained cracks that models such as PAN and DeepLabV3Plus often miss.

The DeepCrack dataset further highlights Context-CrackNet’s strengths, achieving an mIoU of \textbf{0.7401} and Dice Score of \textbf{0.8505}, validating its ability to detect tiny and small crack patterns with high precision. For non-road cracks in datasets like Eugen Miller and Rissbilder, \textit{Context-CrackNet} shows strong adaptability with recalls of \textbf{0.9434} and \textbf{0.8469}, outperforming models that struggle with material and texture variations.

\begin{figure}[!h]
\centering
\includegraphics[width=0.99\textwidth]{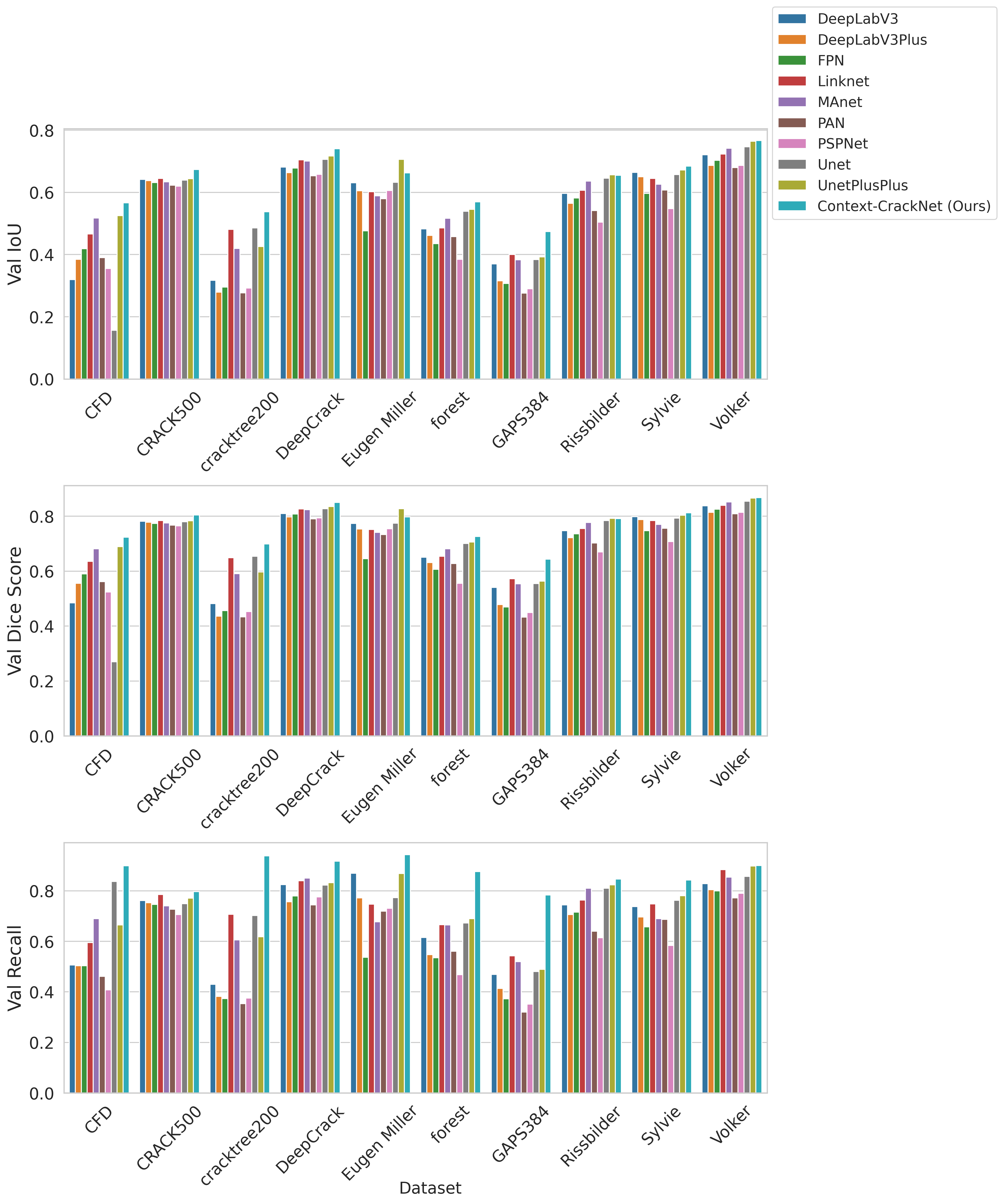}
\caption{Performance comparison of the trained models across the different crack datasets. The top bar chart shows the Validation IoU, the middle bar chart shows the Validation Dice Score, and the bottom bar chart shows the Validation Recall. Higher bars indicate better performance.}
 \label{fig:three_bar_subplots}
\end{figure}

On datasets like GAPs and Forest, where irregular crack features dominate, \textit{Context-CrackNet} consistently achieves superior metrics, confirming its effectiveness in challenging environments. Finally, on Sylvie and Volker, it maintains top performance with mIoUs of \textbf{0.6846} and \textbf{0.7668}, demonstrating its ability to handle varying complexities and environmental conditions. Figure \ref{fig:three_bar_subplots} compares \textit{Context-CrackNet} with other state-of-the-art models across the various crack datasets, highlighting Validation IoU, Dice Score, and Recall.

These results affirm that \textit{Context-CrackNet} effectively addresses the limitations of existing models by reliably detecting tiny and complex cracks across diverse datasets, reinforcing its potential for real-world applications in crack detection and infrastructure monitoring.

\begin{table}[H]
\centering
\caption{Validation Results of \textit{Context-CrackNet} and Other Models Across CFD, CRACK500, CrackTree200, and DeepCrack dataset}
\label{tab:combined_results}
\resizebox{\textwidth}{!}{%
\begin{tabular}{lcccccc}
\hline
\textbf{Dataset} & \textbf{Model} & \textbf{mIoU} & \textbf{Dice (F1) Score} & \textbf{Recall} & \textbf{Precision} \\ \hline
\multirow{10}{*}{CFD} 
& DeepLabV3 & 0.3194 & 0.4842 & 0.5065 & 0.4638 \\
& DeepLabV3Plus & 0.3848 & 0.5558 & 0.5039 & 0.6195 \\
& FPN & 0.4187 & 0.5902 & 0.5033 & 0.7134 \\
& Linknet & 0.4664 & 0.6362 & 0.5951 & 0.6833 \\
& MAnet & 0.5174 & 0.6819 & 0.6901 & 0.6740 \\
& PAN & 0.3904 & 0.5616 & 0.4614 & 0.7173 \\
& PSPNet & 0.3553 & 0.5244 & 0.4084 & \textbf{\textcolor{red}{0.7322}} \\
& Unet & 0.1562 & 0.2703 & 0.8370 & 0.1611 \\
& UnetPlusPlus & 0.5257 & 0.6891 & 0.6653 & 0.7147 \\
& \textbf{Context-CrackNet (ours)} & \textbf{\textcolor{red}{0.5668}} & \textbf{\textcolor{red}{0.7235}} & \textbf{\textcolor{red}{0.8989}} & 0.6054 \\ \hline

\multirow{10}{*}{CRACK500} 
& DeepLabV3 & 0.6420 & 0.7817 & 0.7612 & 0.8035 \\
& DeepLabV3Plus & 0.6376 & 0.7784 & 0.7529 & 0.8060 \\
& FPN & 0.6317 & 0.7740 & 0.7460 & 0.8047 \\
& Linknet & 0.6450 & 0.7839 & 0.7855 & 0.7827 \\
& MAnet & 0.6341 & 0.7757 & 0.7408 & 0.8142 \\
& PAN & 0.6231 & 0.7674 & 0.7271 & 0.8132 \\
& PSPNet & 0.6197 & 0.7649 & 0.7058 & \textbf{\textcolor{red}{0.8356}} \\
& Unet & 0.6397 & 0.7800 & 0.7496 & 0.8133 \\
& UnetPlusPlus & 0.6444 & 0.7835 & 0.7714 & 0.7960 \\
& \textbf{Context-CrackNet (ours)} & \textbf{\textcolor{red}{0.6733}} & \textbf{\textcolor{red}{0.8046}} & \textbf{\textcolor{red}{0.7967}} & 0.8129 \\ \hline

\multirow{10}{*}{CrackTree200} 
& DeepLabV3 & 0.3174 & 0.4819 & 0.4305 & 0.5471 \\
& DeepLabV3Plus & 0.2789 & 0.4361 & 0.3823 & 0.5077 \\
& FPN & 0.2958 & 0.4566 & 0.3736 & 0.5869 \\
& Linknet & 0.4807 & 0.6493 & 0.7066 & 0.6006 \\
& MAnet & 0.4197 & 0.5913 & 0.6063 & 0.5770 \\
& PAN & 0.2765 & 0.4333 & 0.3532 & 0.5602 \\
& PSPNet & 0.2927 & 0.4528 & 0.3757 & 0.5697 \\
& Unet & 0.4857 & 0.6539 & 0.7022 & \textbf{\textcolor{red}{0.6118}} \\
& UnetPlusPlus & 0.4257 & 0.5972 & 0.6179 & 0.5778 \\
& \textbf{Context-CrackNet (ours)} & \textbf{\textcolor{red}{0.5375}} & \textbf{\textcolor{red}{0.6992}} & \textbf{\textcolor{red}{0.9386}} & 0.5571 \\ \hline

\multirow{10}{*}{DeepCrack} 
& DeepLabV3 & 0.6813 & 0.8102 & 0.8246 & 0.7963 \\
& DeepLabV3Plus & 0.6639 & 0.7977 & 0.7570 & \textbf{\textcolor{red}{0.8433}} \\
& FPN & 0.6783 & 0.8081 & 0.7796 & 0.8387 \\
& Linknet & 0.7047 & 0.8267 & 0.8402 & 0.8137 \\
& MAnet & 0.7007 & 0.8238 & 0.8502 & 0.7991 \\
& PAN & 0.6535 & 0.7902 & 0.7444 & 0.8421 \\
& PSPNet & 0.6582 & 0.7936 & 0.7761 & 0.8120 \\
& Unet & 0.7061 & 0.8275 & 0.8231 & 0.8320 \\
& UnetPlusPlus & 0.7168 & 0.8349 & 0.8324 & 0.8374 \\
& \textbf{Context-CrackNet (ours)} & \textbf{\textcolor{red}{0.7401}} & \textbf{\textcolor{red}{0.8505}} & \textbf{\textcolor{red}{0.9175}} & 0.7926 \\ \hline

\end{tabular}%
}
\end{table}

\begin{table}[H]
\centering
\caption{Validation Results of \textit{Context-CrackNet} and Other Models Across Eugen Miller, Forest, GAPs, Rissbilder dataset}
\label{tab:combined_results_2}
\resizebox{\textwidth}{!}{%
\begin{tabular}{lcccccc}
\hline
\textbf{Dataset} & \textbf{Model} & \textbf{mIoU} & \textbf{Dice (F1) Score} &  \textbf{Recall} & \textbf{Precision} \\ \hline

\multirow{10}{*}{Eugen Miller} 
& DeepLabV3 & 0.6312 & 0.7739 & 0.8696 & 0.6971 \\
& DeepLabV3Plus & 0.6051 & 0.7540 & 0.7717 & 0.7371 \\
& FPN & 0.4763 & 0.6452 & 0.5366 & 0.8091 \\
& Linknet & 0.6024 & 0.7519 & 0.7473 & 0.7565 \\
& MAnet & 0.5887 & 0.7411 & 0.6777 & \textbf{\textcolor{red}{0.8176}} \\
& PAN & 0.5794 & 0.7337 & 0.7198 & 0.7482 \\
& PSPNet & 0.6058 & 0.7545 & 0.7311 & 0.7794 \\
& Unet & 0.6323 & 0.7747 & 0.7728 & 0.7767 \\
& UnetPlusPlus & \textbf{\textcolor{red}{0.7060}} & \textbf{\textcolor{red}{0.8277}} & 0.8689 & 0.7902 \\
& \textbf{Context-CrackNet (ours)} & 0.6627 & 0.7971 & \textbf{\textcolor{red}{0.9434}} & 0.6901 \\ \hline

\multirow{10}{*}{Forest} 
& DeepLabV3 & 0.4826 & 0.6510 & 0.6151 & 0.6915 \\
& DeepLabV3Plus & 0.4616 & 0.6316 & 0.5473 & \textbf{\textcolor{red}{0.7465}} \\
& FPN & 0.4356 & 0.6069 & 0.5352 & 0.7007 \\
& Linknet & 0.4857 & 0.6538 & 0.6664 & 0.6417 \\
& MAnet & 0.5170 & 0.6816 & 0.6650 & 0.6990 \\
& PAN & 0.4579 & 0.6281 & 0.5605 & 0.7143 \\
& PSPNet & 0.3848 & 0.5557 & 0.4687 & 0.6825 \\
& Unet & 0.5390 & 0.7005 & 0.6723 & 0.7311 \\
& UnetPlusPlus & 0.5457 & 0.7061 & 0.6896 & 0.7234 \\
& \textbf{Context-CrackNet (ours)} & \textbf{\textcolor{red}{0.5699}} & \textbf{\textcolor{red}{0.7261}} & \textbf{\textcolor{red}{0.8758}} & 0.6201 \\ \hline

\multirow{10}{*}{GAPs} 
& DeepLabV3 & 0.3704 & 0.5405 & 0.4694 & 0.6393 \\
& DeepLabV3Plus & 0.3153 & 0.4785 & 0.4142 & 0.5679 \\
& FPN & 0.3071 & 0.4699 & 0.3729 & 0.6353 \\
& Linknet & 0.4007 & 0.5721 & 0.5430 & 0.6050 \\
& MAnet & 0.3832 & 0.5540 & 0.5196 & 0.5937 \\
& PAN & 0.2758 & 0.4323 & 0.3200 & 0.6668 \\
& PSPNet & 0.2898 & 0.4491 & 0.3518 & 0.6237 \\
& Unet & 0.3837 & 0.5545 & 0.4809 & 0.6548 \\
& UnetPlusPlus & 0.3927 & 0.5637 & 0.4894 & \textbf{\textcolor{red}{0.6669}} \\
& \textbf{Context-CrackNet (ours)} & \textbf{\textcolor{red}{0.4743}} & \textbf{\textcolor{red}{0.6433}} & \textbf{\textcolor{red}{0.7838}} & 0.5456 \\ \hline

\multirow{10}{*}{Rissbilder} 
& DeepLabV3 & 0.5965 & 0.7471 & 0.7442 & 0.7502 \\
& DeepLabV3Plus & 0.5653 & 0.7221 & 0.7057 & 0.7395 \\
& FPN & 0.5819 & 0.7356 & 0.7153 & 0.7572 \\
& Linknet & 0.6068 & 0.7552 & 0.7632 & 0.7476 \\
& MAnet & 0.6365 & 0.7777 & 0.8104 & 0.7477 \\
& PAN & 0.5416 & 0.7024 & 0.6403 & \textbf{\textcolor{red}{0.7783}} \\
& PSPNet & 0.5042 & 0.6700 & 0.6146 & 0.7381 \\
& Unet & 0.6456 & 0.7845 & 0.8105 & 0.7602 \\
& UnetPlusPlus & \textbf{\textcolor{red}{0.6568}} & \textbf{\textcolor{red}{0.7926}} & 0.8233 & 0.7644 \\
& \textbf{Context-CrackNet (ours)} & 0.6553 & 0.7916 & \textbf{\textcolor{red}{0.8469}} & 0.7432 \\ \hline

\end{tabular}%
}
\end{table}

\begin{table}[h!]
\centering
\caption{Validation Results of \textit{Context-CrackNet} and Other Models Across Sylvie and Volker dataset}
\label{tab:combined_results_3}
\resizebox{\textwidth}{!}{%
\begin{tabular}{lcccccc}
\hline
\textbf{Dataset} & \textbf{Model} & \textbf{mIoU} & \textbf{Dice (F1) Score} & \textbf{Recall} & \textbf{Precision} \\ \hline

\multirow{10}{*}{Sylvie} 
& DeepLabV3 & 0.6642 & 0.7982 & 0.7377 & 0.8696 \\
& DeepLabV3Plus & 0.6505 & 0.7882 & 0.6964 & \textbf{\textcolor{red}{0.9079}} \\
& FPN & 0.5966 & 0.7474 & 0.6570 & 0.8666 \\
& Linknet & 0.6449 & 0.7842 & 0.7481 & 0.8238 \\
& MAnet & 0.6263 & 0.7702 & 0.6899 & 0.8718 \\
& PAN & 0.6074 & 0.7558 & 0.6869 & 0.8401 \\
& PSPNet & 0.5481 & 0.7081 & 0.5841 & 0.8989 \\
& Unet & 0.6577 & 0.7935 & 0.7622 & 0.8275 \\
& UnetPlusPlus & 0.6718 & 0.8037 & 0.7804 & 0.8284 \\
& \textbf{Context-CrackNet (ours)} & \textbf{\textcolor{red}{0.6846}} & \textbf{\textcolor{red}{0.8128}} & \textbf{\textcolor{red}{0.8429}} & 0.7847 \\ \hline

\multirow{10}{*}{Volker} 
& DeepLabV3 & 0.7209 & 0.8378 & 0.8280 & 0.8479 \\
& DeepLabV3Plus & 0.6872 & 0.8146 & 0.8046 & 0.8249 \\
& FPN & 0.7028 & 0.8254 & 0.7994 & \textbf{\textcolor{red}{0.8534}} \\
& Linknet & 0.7233 & 0.8394 & 0.8841 & 0.7990 \\
& MAnet & 0.7422 & 0.8520 & 0.8543 & 0.8497 \\
& PAN & 0.6796 & 0.8092 & 0.7720 & 0.8504 \\
& PSPNet & 0.6870 & 0.8144 & 0.7905 & 0.8400 \\
& Unet & 0.7464 & 0.8548 & 0.8574 & 0.8523 \\
& UnetPlusPlus & 0.7641 & 0.8663 & 0.8983 & 0.8365 \\
& \textbf{Context-CrackNet (ours)} & \textbf{\textcolor{red}{0.7668}} & \textbf{\textcolor{red}{0.8680}} & \textbf{\textcolor{red}{0.9002}} & 0.8381 \\ \hline

\end{tabular}%
}
\end{table}

\subsection{Attention Map Visualization from RFEM Modules}
\begin{figure}[H]
\centering
\includegraphics[width=0.71\textwidth]{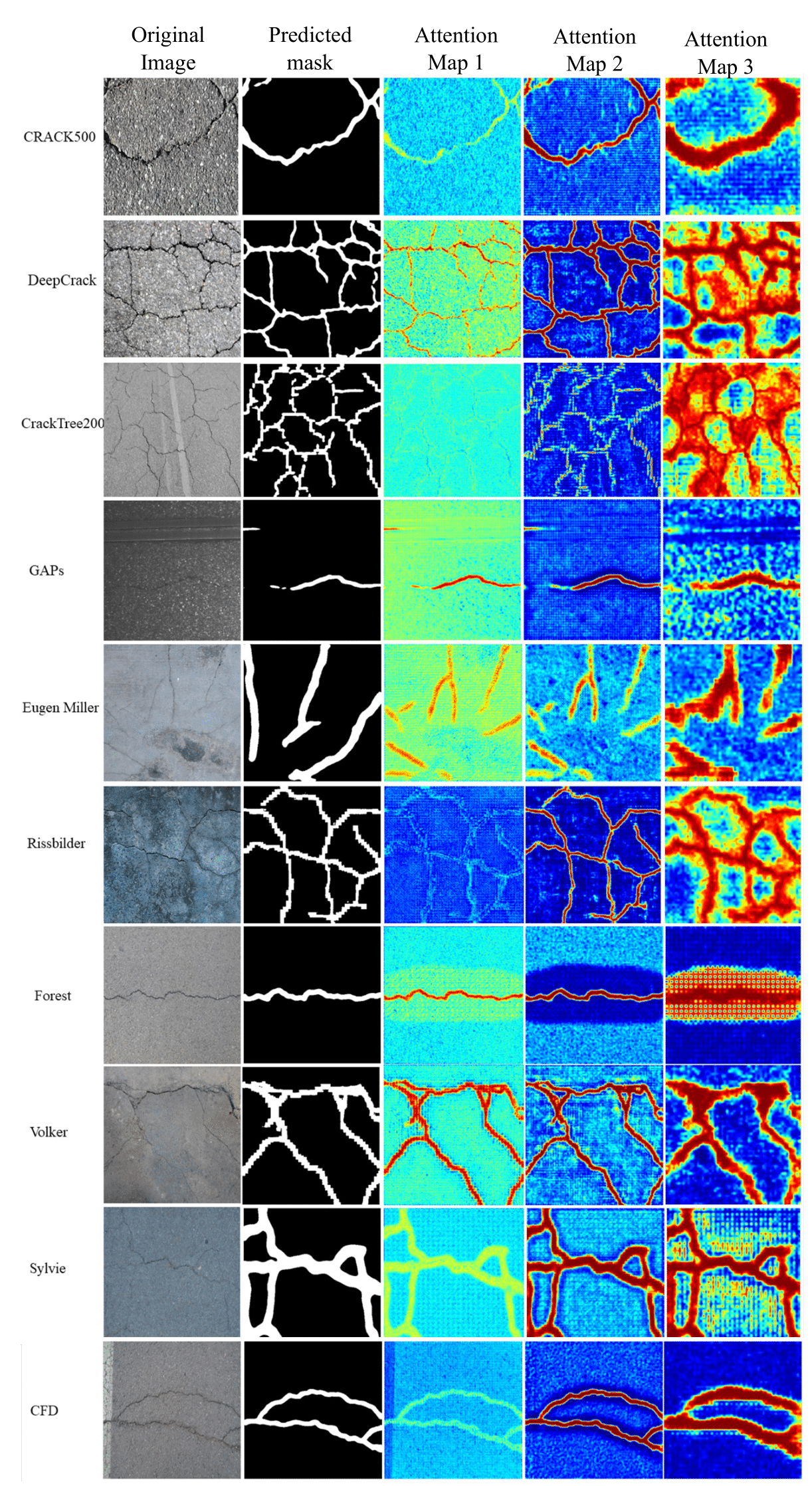}
\caption{Visualization of predicted masks and attention maps from RFEM across various datasets. The attention maps highlight regions of interest contributing to the segmentation predictions from \textit{Context-CrackNet}}
 \label{fig:attention_maps}
\end{figure}

The attention maps in Figure \ref{fig:attention_maps} illustrate the regions of the input images that the RFEM modules in \textit{Context-CrackNet} focuses on during segmentation. These visualizations reveal that the model effectively identifies and emphasizes critical areas of distress across the different datasets. For example, in datasets like CRACK500 and DeepCrack, the attention maps distinctly capture intricate crack patterns, demonstrating the model's ability to localize fine-grained details. In complex cases like Rissbilder and Eugen Miller, the attention maps prioritize regions with subtle texture variations, ensuring accurate predictions even in challenging conditions. By highlighting relevant features, the attention maps provide interpretability to the model’s decisions and validate its capability to generalize across datasets with varying characteristics. This insight is crucial for understanding how the model adapts to different pavement distress types. 

\subsection{Ablation studies}
To thoroughly analyze the contributions of the RFEM and CAGM modules, we conducted ablation experiments by selectively enabling or disabling these modules in the proposed architecture. These experiments allow us to quantify the individual and combined effects of RFEM and CAGM on segmentation performance. The results are summarized in Table \ref{tab:ablation_study}.

\begin{table}[H]
\centering
\caption{Ablation study results for RFEM and CAGM. Metrics include validation IoU, Dice Score, Precision, and Recall.}
\label{tab:ablation_study}
\begin{tabular}{lcccc}
\hline
\textbf{Configuration}      & \textbf{mIoU} & \textbf{Dice Score} & \textbf{Precision} & \textbf{Recall} \\ \hline
Baseline & 0.4259           & 0.5929                  & 0.5481                 & 0.6691              \\
RFEM Only              & 0.4355           & 0.6057                  & 0.5439                 & 0.6990              \\
CAGM Only              & 0.4263           & 0.5954                  & 0.5392                 & 0.6827              \\
RFEM + CAGM     & \textbf{0.4743}  & \textbf{0.6433}         & \textbf{0.5456}        & \textbf{0.7838}     \\ \hline
\end{tabular}
\end{table}

\paragraph{\textbf{Analysis and Discussion}}
The baseline model, without the RFEM and CAGM modules, achieved a validation IoU of 0.4259 and a Dice Score of 0.5929, demonstrating limited capability in capturing both local and global context. Adding RFEM alone improved the IoU to 0.4355 and the Dice Score to 0.6057, indicating that RFEM effectively enhances the model's ability to focus on critical local regions, especially fine-grained crack details. This is further reflected in the increase in recall from 0.6691 to 0.6990, as RFEM enables the model to identify more instances of distress.

When CAGM was included without RFEM, the IoU and Dice Score showed minimal improvements (0.4263 and 0.5954, respectively). While CAGM provides global context by capturing broader spatial dependencies, its contribution is less pronounced when local refinement (via RFEM) is absent. However, recall improved to 0.6827, suggesting that CAGM aids in generalizing to larger contextual regions, albeit at the expense of precision.

The model performed best when both RFEM and CAGM were included, achieving an IoU of 0.4743 and a Dice Score of 0.6433. The significant boost in recall to 0.7838 highlights the complementary roles of RFEM and CAGM. RFEM sharpens the model’s focus on localized crack patterns, while CAGM enriches the global context, leading to better overall segmentation. Interestingly, precision did not increase significantly with the inclusion of both modules, remaining relatively stable. This suggests that while the model identifies distress regions more effectively, some misclassifications persist, warranting further refinement.

\subsection{Error Cases and Failure Scenarios}
\begin{figure}[H]
\centering
\includegraphics[width=0.8\textwidth]{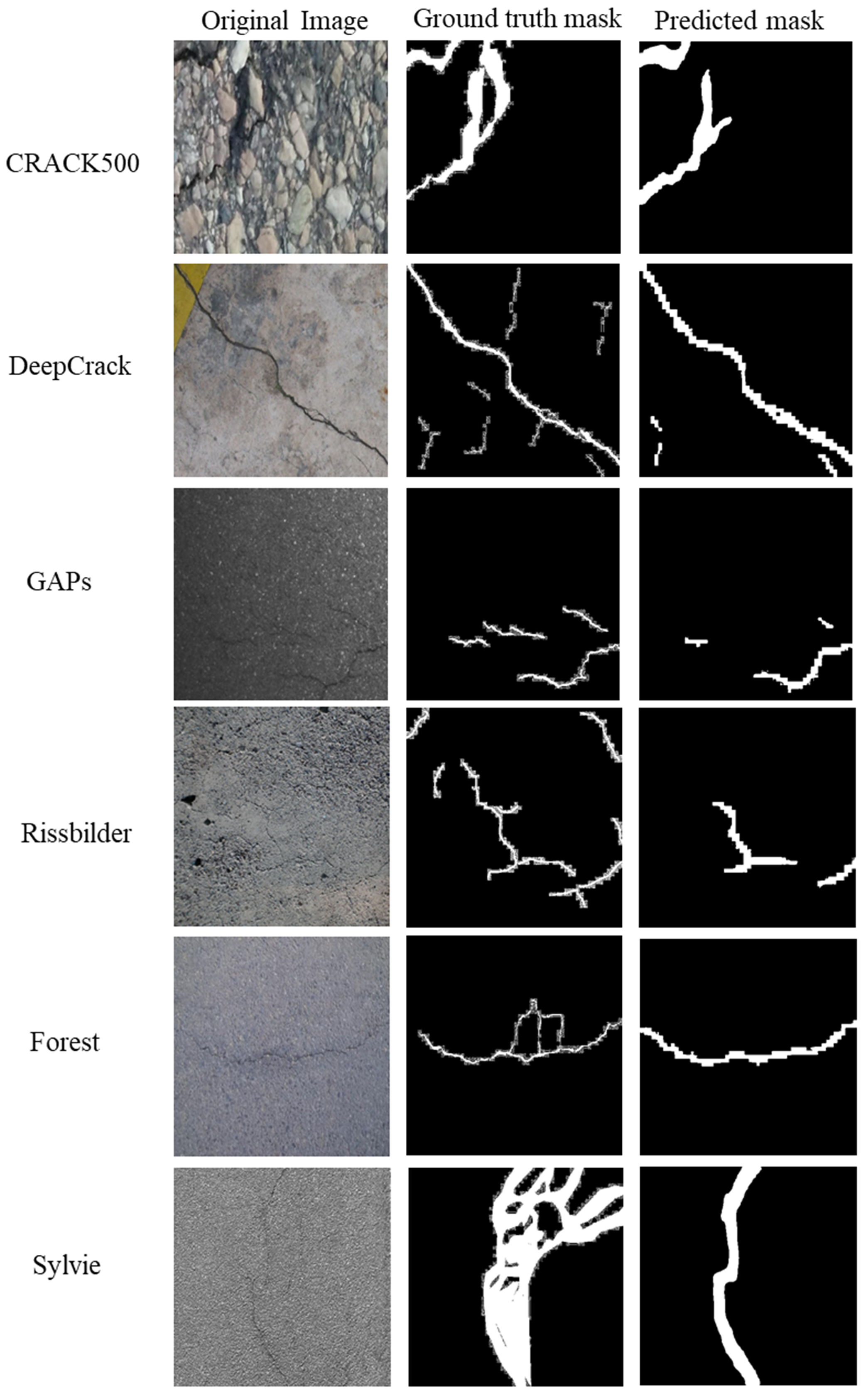}
\caption{Failure cases from Context-CrackNet across various crack datasets, showing discrepancies between ground truth and predicted masks.}
 \label{fig:failure-cases}
\end{figure}

Figure \ref{fig:failure-cases} highlights several failure scenarios from \textit{Context-CrackNet} across some of the crack datasets showing errors between ground truth masks and predicted masks. These failure scenarios often stem from inherent challenges in the datasets, including low contrast between the cracks and their surroundings, and noise or texture inconsistencies in the images. For instance, in the CRACK500 dataset, the network struggled to differentiate between closely spaced cracks and surrounding noise, leading to incomplete or missed segments. Similarly, in the DeepCrack dataset, cracks exhibiting faint textures or irregular patterns were either under-segmented or omitted, reflecting the difficulty in capturing fine-grained structures under varying lighting conditions.

In datasets like Rissbilder and Forest, the model’s predictions were influenced by the complex background textures that mimic crack patterns. This suggests that the network may occasionally misinterpret irrelevant features as cracks, especially when contrast is minimal. The GAPs dataset, with its smooth and uniform background, exposed the network's sensitivity to subtle intensity variations, resulting in missed detections for faint cracks. Additionally, the Sylvie dataset presented a unique challenge where abrupt light intensity transitions and non-crack structures confused the model, leading to segmentation errors. 

\subsection{Model Complexity Analysis}
Computational complexity is crucial for real-time applications like pavement distress detection and monitoring. Table \ref{tab:model_complexity} compares the various models based on parameters, inference time, and GFLOPs.\textit{Context-CrackNet}, with 82.05M parameters and 243.78 GFLOPs, achieves a competitive inference time of 15.63 ms. While it has a higher parameter count than other lightweight models like FPN (26.12M, 6.35 ms) and DeepLabV3Plus (26.68M, 6.72 ms), it offers superior capacity to handle complex spatial relationships crucial for crack segmentation.

Compared to UNetPlusPlus, which has higher FLOPs (352.68 GFLOPs) and slower inference (24.44 ms), \textit{Context-CrackNet} strikes a better balance between performance and efficiency (see Tables \ref{tab:combined_results}, \ref{tab:combined_results_2}, and \ref{tab:combined_results_3}). This balance makes it feasible for real-time deployment and scalable for large-scale pavement monitoring, where both precision and responsiveness are critical. Context-CrackNet's architecture effectively addresses the challenges of robustness and generalization across diverse pavement surfaces, making it well-suited for practical applications.
\begin{table}[!ht]
\centering
\caption{Comparison of model complexity metrics across different architectures. 
Metrics include: 
(1) The number of parameters (in millions), 
(2) The average inference time (in ms) to process a single 448$\times$448 image, 
and (3) GFLOPs (the number of floating-point operations, in billions, required for a single forward pass).}
\label{tab:model_complexity}
\begin{tabular}{lccc}
\hline
\textbf{Model} & \makecell{\textbf{Parameters} \\ \textbf{(M)}} & \makecell{\textbf{Inference Time (ms)} \\ \textbf{(448 x 448 Image)}} & \textbf{GFLOPs} \\
\hline
DeepLabV3          & 39.63  & 17.19 & 251.28 \\
DeepLabV3Plus         & 26.68  & 6.72  & 56.42  \\
FPN                & 26.12  & 6.35  & 48.08  \\
LinkNet            & 31.18  & 6.85  & 66.06  \\
MANet              & 147.44 & 12.85 & 114.48 \\
PAN                & 24.26  & 7.01  & 53.46  \\
PSPNet             & 24.26  & 3.09  & 18.14  \\
UNet               & 32.52  & 7.67  & 65.60  \\
UNetPlusPlus             & 48.99  & 24.44 & 352.68 \\
\textbf{Context-CrackNet} & 82.05  & 15.63 & 243.78 \\
\hline
\end{tabular}
\end{table}

\section{Conclusion}
This study introduces Context-CrackNet, a novel architecture designed to address the persistent challenges of precise segmentation of tiny and subtle pavement cracks. By integrating the Region-Focused Enhancement Module (RFEM) and the Context-Aware Global Module (CAGM), the model demonstrates the ability to capture fine-grained crack details and employ global contextual information for robust segmentation. Extensive experiments across ten publicly available datasets highlight its superior performance over state-of-the-art models, particularly in handling multi-scale crack patterns and challenging real-world conditions.

The results confirm that Context-CrackNet effectively balances segmentation accuracy and computational efficiency, making it suitable for real-time applications. The analysis of error cases underscores the complexities associated with varying lighting, noise, and surface textures in pavement datasets, suggesting avenues for further model refinement. Additionally, the comprehensive model complexity analysis reveals that while Context-CrackNet entails higher computational demands than some lightweight models, it achieves a favorable trade-off by providing significantly enhanced segmentation performance. This balance ensures the model's feasibility for large-scale deployment in pavement monitoring systems, where precision and scalability are paramount.

The findings of this research have broad implications for automated pavement distress detection and monitoring. By enabling early detection of tiny cracks, Context-CrackNet contributes to preventive maintenance strategies that can reduce repair costs and improve the longevity of road networks. Furthermore, the proposed architecture sets a foundation for developing context-aware frameworks applicable to other domains requiring fine-grained segmentation under resource-constrained environments.

Future research can explore several promising directions. Incorporating advanced data augmentation techniques and domain adaptation methods may further enhance the model's generalizability to unseen datasets. Additionally, developing lightweight attention mechanisms or model compression techniques could reduce computational overhead without compromising accuracy, thereby broadening its applicability to edge devices. Finally, extending the framework to multi-class segmentation tasks or integrating it with predictive maintenance systems could unlock new opportunities for infrastructure management.
\bibliographystyle{unsrt}
\bibliography{export}
\end{document}